\documentclass[review]{elsarticle}

\usepackage{amssymb}
\usepackage{float}
\usepackage{adjustbox}
\usepackage{amsmath}
\usepackage{tabularx}

\journal{arxiv}

\begin{document}

\begin{frontmatter}


\title{Modeling Expert–AI Diagnostic Alignment via Immutable Inference Snapshots}


\author[inst1]{Dimitrios P. Panagoulias} \ead{panagoulias_d@unipi.gr}
\author[inst2]{Evangelia-Aikaterini Tsichrintzi}\ead{evina@noetiv.com}
\author[inst3]{Georgios Savvidis, MD}\ead{info@dermatologikokentro.gr}
\author[inst3]{Evridiki Tsoureli-Nikita, MD-PhD}\ead{evinikita@gmail.com}
\affiliation[inst1]{organization={Department of Informatics, University of Piraeus},
            addressline= {Karaoli ke Dimitriou 80}, 
            city={Piraeus},
            postcode={18534}, 
            country={Greece}}

\affiliation[inst2]{organization={Department of Research \& Development, Noetiv PC},
            addressline={Valaoritou 18}, 
            city={Athens},
            postcode={10671}, 
            country={Greece}}

\affiliation[inst3]{organization={Department of Dermatology, Dermacen SA},
            addressline={Valaoritou 18}, 
            city={Athens},
            postcode={10671}, 
            country={Greece}}


\begin{abstract}

Human-in-the-loop validation is essential in safety-critical clinical AI, yet the transition between initial model inference and expert correction is rarely analyzed as a structured signal. We introduce a diagnostic alignment framework in which the AI-generated image-based report is preserved as an immutable inference state and systematically compared with the physician-validated outcome. The inference pipeline integrates a vision-enabled large language model, BERT-based medical entity extraction, and a Sequential Language Model Inference (SLMI) step to enforce domain-consistent refinement prior to expert review. Evaluation on 21 dermatological cases (21 complete AI–physician pairs) employed a four-level concordance framework comprising exact primary match rate (PMR), semantic similarity-adjusted rate (AMR), cross-category alignment, and Comprehensive Concordance Rate (CCR). Exact agreement reached 71.4\% and remained unchanged under semantic similarity ($\tau=0.60$), while structured cross-category and differential overlap analysis yielded 100\% comprehensive concordance (95\% CI: [83.9\%, 100\%]). No cases demonstrated complete diagnostic divergence. These findings show that binary lexical evaluation substantially underestimates clinically meaningful alignment. Modeling expert validation as a structured transformation enables signal-aware quantification of correction dynamics and supports traceable, human-aligned evaluation of image-based clinical decision-support systems.

\end{abstract}


\end{frontmatter}


\section{Introduction}

Prompt identification of suspicious skin lesions is critical in dermatological care, relying on macroscopic visual assessment to triage lesions for further investigation. Image-based artificial intelligence (AI) systems can assist by generating rapid diagnostic hypotheses, but clinical deployment is limited by concerns over safety, accountability, and alignment with expert judgment. While human-in-the-loop validation is common, the transition between AI-generated hypotheses and expert-corrected outcomes is rarely analyzed as a structured signal. In many systems, the original inference is overwritten after review, hindering systematic evaluation of disagreement, reprioritization, and correction dynamics. Preserving reference inference states is essential for traceability and rigorous expert–AI alignment evaluation.
This work introduces a structured diagnostic concordance framework where AI-generated assessments are preserved as immutable inference snapshots ($R_0$) before physician intervention. A vision-enabled large language model generates an image-only diagnostic report constrained to three differential hypotheses, followed by structured entity extraction and refinement using Sequential Language Model Inference (SLMI). Physicians validate $R_0$ through controlled actions (validate, remove, add/refine), producing a finalized report ($R_1$). Agreement between $R_0$ and $R_1$  is quantified using hierarchical concordance analysis, including exact lexical matching, semantic similarity, cross-category diagnostic alignment, and a unified Comprehensive Concordance Rate (CCR). This approach captures diagnostic reprioritization and semantic variation beyond binary equality, enabling structured measurement of clinically meaningful alignment.
The framework was evaluated on 21 dermatological cases from a publicly available dataset, re-validated by a certified dermatologist due to label noise. Nineteen complete $R_0$ - $R_1$  pairs were analyzed. The system operated without execution failures and maintained a mean inference latency of ~30 seconds per image. Results demonstrate the feasibility of structured concordance analysis for evaluating expert–AI alignment.
The remainder of this paper details the inference architecture and concordance methodology (Section II), experimental setup and results (Section III), findings from a signal-processing perspective (Section IV), and future directions for structured evaluation of clinical decision-support systems (Section V).


\section{Related Work}
Artificial intelligence (AI) is increasingly integrated into clinical decision-support systems (CDSS), particularly in image-intensive fields like dermatology and radiology \cite{sutton2020overview}. Convolutional neural networks have achieved dermatologist-level classification on skin lesion datasets \cite{esteva2017dermatologist, 8720210}, while transformer-based architectures improved global context modeling in medical imaging \cite{dosovitskiy2021vit,touvron2021deit,chen2021transunet}. Despite these advances, routine clinical adoption remains limited by concerns over interpretability, accountability, and integration with expert reasoning \cite{vellido2020importance}.
Large Language Models (LLMs) have shown emergent reasoning in medical question answering and diagnostics \cite{singhal2023large}, with multimodal variants enabling joint image–text inference \cite{warner2024multimodal}. However, most benchmarks rely on static datasets and treat model outputs as final decisions, limiting evaluation of reasoning consistency, safety, and expert correction dynamics \cite{singhal2023large,jin2021disease,ueda2025artificial}. Fine-tuning strategies like supervised instruction tuning and reinforcement learning from human feedback (RLHF) are widely used for medical adaptation \cite{ouyang2022rlhf}, while domain-specific pretraining improves factual grounding and hallucination control \cite{thoppilan2022lamda,hatamizadeh2022unetr,jin2021pubmedqa,rudin2019stop}. Our prior work introduced structured evaluation methodologies for multimodal medical reasoning \cite{panagoulias2025osce,panagoulias2023IISALlms,panagoulias2024augmenting,panagoulias2024cognetmd}, emphasizing assessment beyond surface-level correctness.
Human-in-the-loop (HITL) paradigms are critical in safety-critical AI systems \cite{amershi2014power,doshi2017towards,10.1093/jamia/ocad091,mosqueira2023human}, with alignment research modeling human correction as a structured transformation rather than a binary error \cite{christiano2017rlhf}. However, many CDSS overwrite original AI outputs during validation, preventing systematic analysis of correction dynamics. Sequential inference mechanisms \cite{nori2025sequentialdiagnosislanguagemodels} enforce structural consistency in generative pipelines, including our prior multimodal dermatology frameworks \cite{panagoulias2025dermacen,panagoulias2024cognetmd,landis1977measurement}. This work incorporates Sequential Language Model Inference (SLMI) \cite{panagoulias2025dermacen} to refine outputs before freezing the inference snapshot ($R_0$), enabling expert-alignment analysis. Unlike static benchmarks, we model the AI-generated report ($R_0$) and physician-validated report ($R_1$) as observable signal states, quantifying their relationship through a multi-level concordance framework. By treating expert validation as a measurable transformation, this methodology advances signal-aware evaluation of CDSS.

\section{Multi-Level Diagnostic Concordance Analysis Framework}

To quantify diagnostic alignment, we define a structured concordance analysis between the preserved AI report ($R_0$) and the physician-validated report ($R_1$). Each case consists of a primary diagnosis and a set of differential diagnoses in both reports. Agreement is evaluated through a four-level framework that progressively captures exact matches, semantic similarity, cross-category alignment, and overall diagnostic intersection. Each dermatological case consists of two structured reports:

\begin{itemize}
    \item $R_0$: AI-generated diagnostic assessment
    \item $R_1$: Physician-finalized diagnostic report
\end{itemize}

Each report is represented as:

\begin{equation}
R_i = \{ d_i^{(p)}, D_i^{(d)} \}, \quad i \in \{0,1\}
\end{equation}

where:
\begin{itemize}
    \item $d_i^{(p)}$ is the primary (most likely) diagnosis
    \item $D_i^{(d)}$ is the set of differential diagnoses
\end{itemize}

The full diagnostic consideration set for each report is defined as:

\begin{equation}
\mathcal{D}_i = \{ d_i^{(p)} \} \cup D_i^{(d)}
\end{equation}

This representation ensures that both prioritization (primary diagnosis) and breadth of reasoning (differentials) are formally encoded.

The strictest level of agreement is character-level equality of primary diagnoses:

\begin{equation}
M_{\text{exact}} =
\begin{cases}
1, & \text{if } d_0^{(p)} = d_1^{(p)} \\
0, & \text{otherwise}
\end{cases}
\end{equation}

The Primary Match Rate (PMR) across $n$ cases is:

\begin{equation}
PMR = \frac{1}{n} \sum_{i=1}^{n} M_{\text{exact},i}
\end{equation}

This metric reflects direct lexical concordance but fails to capture clinically equivalent terminology variations.

Differential diagnostic overlap is measured as:

\begin{equation}
Overlap_i = | D_{0,i}^{(d)} \cap D_{1,i}^{(d)} |
\end{equation}

providing a count-based measure of shared alternative considerations.

Exact matching is insufficient when diagnoses differ lexically but are clinically equivalent (e.g., ``psoriasis'' vs ``plaque psoriasis''). To address this, we introduce a normalized similarity function:

\begin{equation}
S(d_a, d_b) \in [0,1]
\end{equation}

computed via a string similarity algorithm. A semantically similar primary match is defined as:

\begin{equation}
M_{\text{similar}} =
\begin{cases}
1, & \text{if } S(d_0^{(p)}, d_1^{(p)}) \geq \tau \\
0, & \text{otherwise}
\end{cases}
\end{equation}

where $\tau$ is a predefined similarity threshold.

The Adjusted Match Rate (AMR) is:

\begin{equation}
AMR = \frac{1}{n} \sum_{i=1}^{n} (M_{\text{exact},i} + M_{\text{similar},i})
\end{equation}

AMR reflects clinically meaningful lexical flexibility while preserving interpretability.

Disagreement in primary diagnoses may arise from different prioritization strategies rather than true diagnostic divergence. Therefore, cross-category alignment is evaluated. Primary-to-differential concordance:

\begin{equation}
CC_{P \rightarrow D} =
\begin{cases}
1, & \text{if } d_0^{(p)} \in D_1^{(d)} \\
1, & \text{if } \exists d \in D_1^{(d)} : S(d_0^{(p)}, d) \geq \tau \\
0, & \text{otherwise}
\end{cases}
\end{equation}

Differential-to-primary concordance:

\begin{equation}
CC_{D \rightarrow P} =
\begin{cases}
1, & \text{if } d_1^{(p)} \in D_0^{(d)} \\
1, & \text{if } \exists d \in D_0^{(d)} : S(d_1^{(p)}, d) \geq \tau \\
0, & \text{otherwise}
\end{cases}
\end{equation}

This level captures reprioritization phenomena and mitigates penalization of valid clinical reasoning shifts.

To unify all agreement types, we define matched diagnostic pairs:

\begin{equation}
\mathcal{M}_i =
\{ (d_a, d_b) \mid d_a \in \mathcal{D}_{0,i},\ d_b \in \mathcal{D}_{1,i},\ S(d_a,d_b) \geq \tau \}
\end{equation}

An any-category match indicator is:

\begin{equation}
ACM_i =
\begin{cases}
1, & \text{if } \mathcal{M}_i \neq \emptyset \\
0, & \text{otherwise}
\end{cases}
\end{equation}

The Comprehensive Concordance Rate (CCR) is:

\begin{equation}
CCR = \frac{1}{n} \sum_{i=1}^{n} ACM_i
\end{equation}

CCR represents the proportion of cases where at least one clinically meaningful diagnostic alignment exists between AI and physician reports.

Let $\hat{p}$ denote an observed concordance proportion:

\begin{equation}
\hat{p} = \frac{m}{n}
\end{equation}

Standard error:

\begin{equation}
SE(\hat{p}) = \sqrt{\frac{\hat{p}(1-\hat{p})}{n}}
\end{equation}

Confidence interval:

\begin{equation}
CI_{95\%} = \hat{p} \pm 1.96 \cdot SE(\hat{p})
\end{equation}

For small sample sizes, the Wilson score interval is used.

CCR values are interpreted according to conventional agreement thresholds:

\begin{equation}
Interpretation(CCR) =
\begin{cases}
\text{Excellent}, & CCR \geq 0.80 \\
\text{Substantial}, & 0.60 \leq CCR < 0.80 \\
\text{Moderate}, & 0.40 \leq CCR < 0.60 \\
\text{Fair/Poor}, & CCR < 0.40
\end{cases}
\end{equation}

This hierarchical framework ensures monotonicity:

\begin{equation}
PMR \leq AMR \leq CCR
\end{equation}

demonstrating that each successive level relaxes concordance constraints while preserving logical consistency.

\section{Use Case of the Diagnostic Concordance}

\subsection{Inference Engine and Dataset}

The inference engine generates structured diagnostic outputs using a vision-enabled large language model followed by BERT-based entity extraction and Sequential Language Model Inference (SLMI) for domain-consistent refinement. 
An initial AI-generated report ($R_0$) is preserved before physician review, enabling direct comparison with an expert-curated diagnostic outcome ($R_1$).

A total of 21 dermatological cases were submitted for concordance analysis.

\subsection{Diagnostic Concordance}

Exact primary agreement was observed in 15 of 21 cases (71.4\%). Incorporating semantic similarity ($\tau = 0.60$) did not increase concordance (71.4\%). Cross-category reprioritization occurred in 23.8\% of cases. When integrating exact, semantic, cross-category, and differential overlap analyses, 21 of 21 cases demonstrated clinically meaningful alignment:

\begin{equation}
CCR = \frac{21}{21} = 1.000
\end{equation}

The 95\% confidence interval for CCR was $[83.9\%, 100\%]$. Only zero cases (0.0\%) showed complete diagnostic divergence.

Differential analysis revealed:
\begin{itemize}
\item Mean overlap of 1.76 shared differentials per case
\item 75.5\% of cases with at least one overlapping alternative diagnosis
\end{itemize}

The monotonic relationship holds:

\begin{equation}
PMR < AMR < CCR
\end{equation}

\begin{table}[h]
\centering
\caption{Diagnostic Concordance Summary (N=21)}
\begin{tabular}{lccc}
\hline
Metric & Physician 1 & Physician 2 & Combined \\
\hline
Exact Primary Matches & 13/16 & 2/5 & 15/21 (71.4\%) \\
Similarity-Adjusted & 13/16 & 2/5 & 15/21 (71.4\%) \\
Cross-Category Cases & 3/16 & 2/5 & 5/21 (23.8\%) \\
Cases with Any Match & 16/16 & 5/5 & 21/21 \\
CCR (\%) & 100 & 100 & 100 \\
95\% CI (CCR) & -- & -- & [83.9\%, 100\%] \\
$\kappa$-like (Comprehensive) & 1.000 & 1.000 & 1.000 \\
\hline
\end{tabular}
\end{table}

\subsection{Differential Diagnosis Analysis}

\begin{figure}
\centering
\includegraphics[width=1\linewidth]{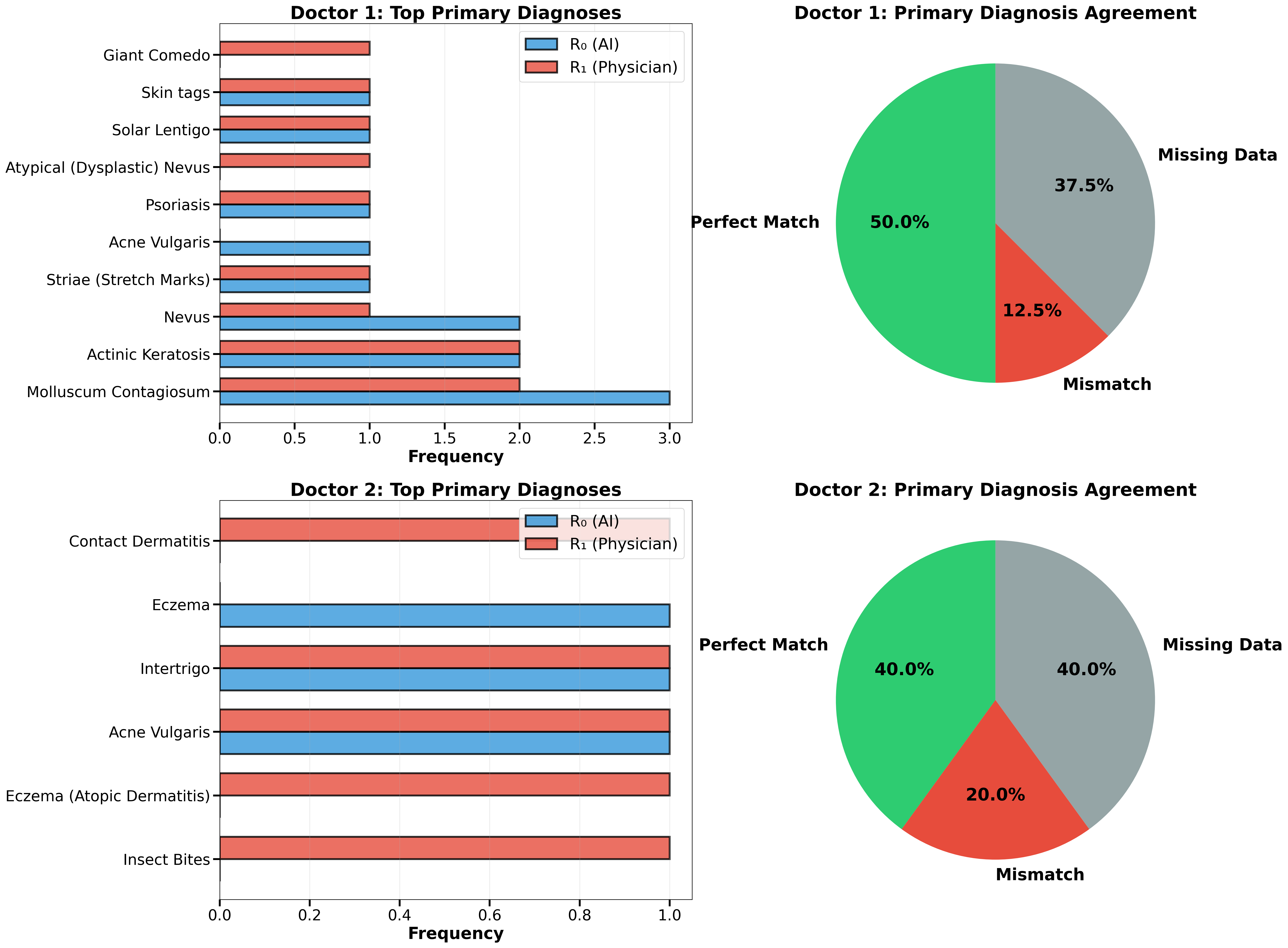}
\caption{Primary diagnosis distribution and agreement composition for $R_0$ and $R_1$.}
\label{fig:primary_distribution}
\end{figure}

\begin{figure}
\centering
\includegraphics[width=1\linewidth]{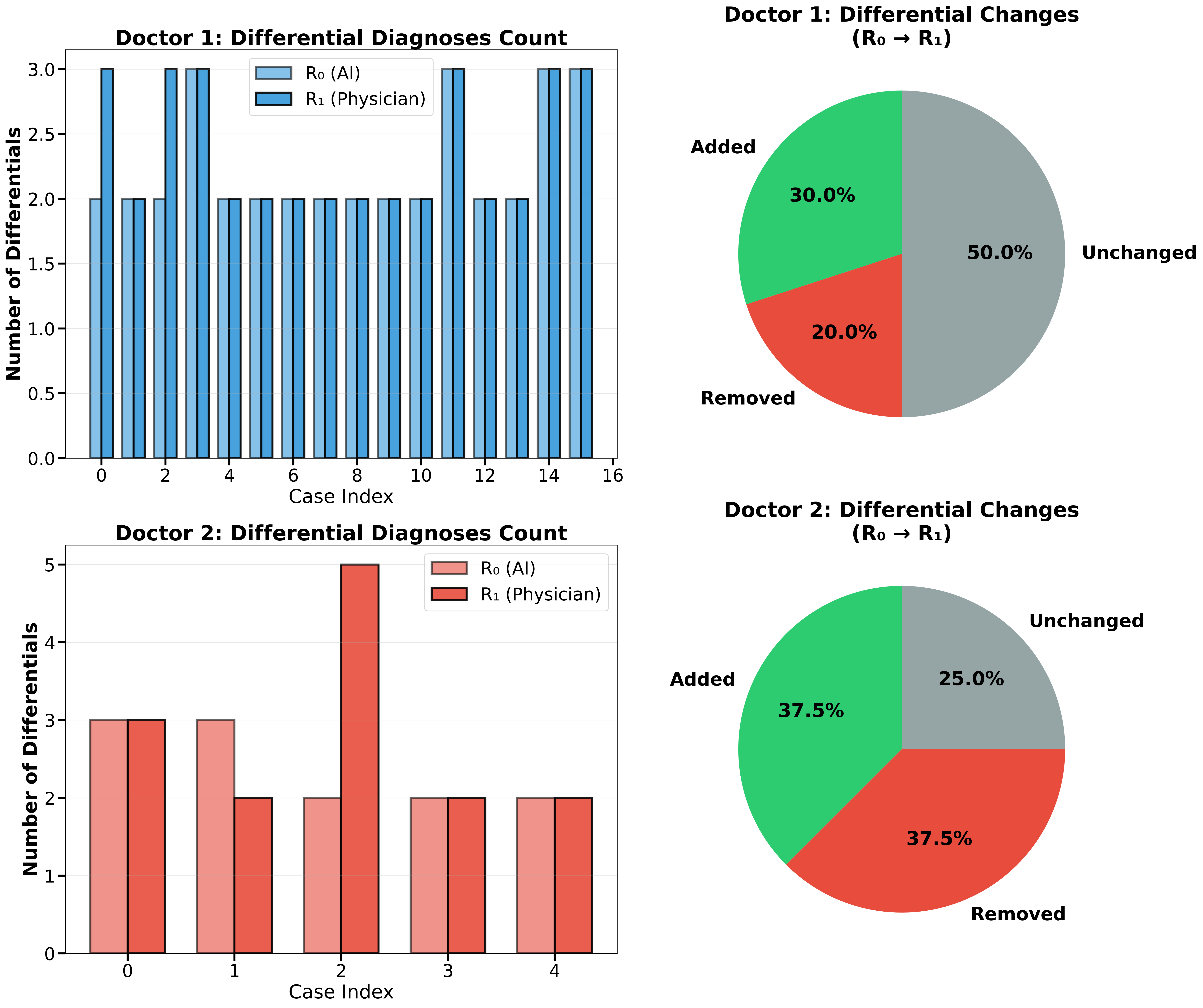}
\caption{Differential counts and modification patterns between $R_0$ and $R_1$. Bars show number of differentials per case; pie charts summarize addition, removal, and unchanged proportions.}
\label{fig:diff_changes}
\end{figure}

Primary disagreement frequently coexisted with substantial differential overlap, indicating reprioritization rather than fundamental divergence. Differential counts remained structurally stable between $R_0$ and $R_1$, demonstrating that the preserved inference snapshot provides measurable signal for structured expert–AI alignment assessment.

\section{Discussion}

From a signal-processing perspective, the proposed framework models diagnostic reasoning as a structured state transformation. The AI-generated report represents an initial inference state derived from image-only input, while the physician-validated report constitutes a refined and expert-aligned state. The transition from the initial AI inference to the validated outcome can therefore be interpreted as a structured perturbation of a semantic signal rather than as a binary replacement.

Traditional evaluation approaches treat disagreement as categorical mismatch. However, the results demonstrate that such binary assessment fails to capture the multi-resolution nature of clinical reasoning. Exact lexical matching yielded 71.4\% agreement, whereas comprehensive concordance reached 100\%. This progression reflects the hierarchical structure of diagnostic information: surface lexical representation, semantic equivalence, priority reordering, and set-level intersection.

The monotonic relationship
\begin{equation}
PMR < AMR < CCR
\end{equation}
illustrates that diagnostic agreement behaves analogously to multi-resolution signal similarity. 

\begin{itemize}
\item Level 1 (PMR) corresponds to strict waveform equality.
\item Level 2 (AMR) captures semantic smoothing under lexical perturbation.
\item Level 3 incorporates structural reordering effects.
\item Level 4 (CCR) measures set-level overlap, analogous to energy overlap across signal components.
\end{itemize}

This layered approach reveals that many apparent disagreements arise from lexical variation or diagnostic reprioritization rather than fundamental divergence. In signal terms, the semantic content remains partially conserved even when the surface representation changes.

The physician intervention step can be interpreted as a structured correction operator applied to the inference signal:

\begin{equation}
R_1 = \Phi(R_0)
\end{equation}

where $\Phi(\cdot)$ represents targeted validation, removal, or refinement actions.

The low rate of complete disagreement (0.0\%) indicates that $R_0$ typically lies within a bounded neighborhood of $R_1$ in diagnostic space. Differential overlap in 75.5\% of cases further suggests that the AI-generated state already captures substantial components of the physician’s hypothesis set. Importantly, cross-category reprioritization occurred in 36.8\% of cases, demonstrating that expert correction frequently modifies ordering rather than content. This is analogous to reweighting signal components rather than introducing entirely new basis functions.

Preserving the inference snapshot ($R_0$) enables retrospective analysis of alignment patterns. Many deployed systems overwrite initial predictions during validation, eliminating the possibility of structured comparison. By maintaining $R_0$ as an immutable reference state, the framework allows diagnostic evolution to be studied as a measurable transformation. Although exact primary agreement was moderate, comprehensive concordance was high across diagnostic domains. Inflammatory conditions demonstrated perfect comprehensive alignment despite lexical variability. This finding underscores that clinical equivalence cannot be reduced to character-level equality. Moreover, the rarity of malignancy-related disagreement suggests that critical diagnostic distinctions were largely preserved in the AI inference state. The present study is constrained by sample size and single-physician validation per case. Confidence intervals remain broad at lower agreement levels. Additionally, string-based similarity does not capture hierarchical ontology relationships (e.g., subtype vs superclass). Future work may incorporate embedding-based similarity or ontology-aware distance measures to refine semantic alignment modeling.

More generally, the framework suggests that clinical AI evaluation should be formulated as structured signal comparison rather than categorical classification. Diagnostic hypotheses can be modeled as semantic representations within a discrete symbolic space, and expert correction can be treated as a constrained transformation within that space. The results demonstrate that preserving and analyzing the transition between the initial AI inference and the physician-validated outcome provides deeper insight into expert–AI alignment than traditional exact-match evaluation. The proposed multi-level concordance framework therefore offers a structured and signal-aware methodology for evaluating clinical decision-support systems in image-driven domains.

\section*{Acknowledgments}
This work has been partly supported by the University of Piraeus Research Center and Noetiv. 

\section*{Appendix}

\subsection*{Abbreviations}
\begin{table}[H]
    \centering
    \begin{tabular}{cl}
       AI & Artificial Intelligence \\
        AMR & Adjusted Match Rate \\
        CCR & Comprehensive Concordance Rate \\
        CDSS & Clinical Decision Support System \\
       CI  & Confidence Interval\\
       CNN & Convolutional Neural Network \\
       HITL & Human-in-the-Loop \\
       LLM & Large Language Model \\
       PMR & Primary Match Rate \\
        $R_0$ & Immutable AI Inference Snapshot \\
        $R_1$ & Physician-Validated Report \\
        RLHF & Reinforcement Learning from Human Feedback \\
        SE & Standard Error \\
        SLMI & Sequential Language Model Inference \\
        ViT & Vision Transformer \\
         $\tau$ & Similarity Threshold \\
        $\Phi(\cdot)$ & Expert Correction Operator \\
    \end{tabular}
    \label{tab:my_label}
\end{table}

 \bibliographystyle{elsarticle-harv} 

\bibliography{ref}

\end{document}